\documentclass[conference]{IEEEtran}
\IEEEoverridecommandlockouts
\usepackage{booktabs}
\usepackage{cite}
\usepackage{amsmath,amssymb,amsfonts}
\usepackage{algorithmic}
\usepackage{graphicx}
\usepackage{textcomp}
\usepackage{xcolor}
\def\BibTeX{{\rm B\kern-.05em{\sc i\kern-.025em b}\kern-.08em
    T\kern-.1667em\lower.7ex\hbox{E}\kern-.125emX}}
\begin{document}

\title{A Concise Review of Hallucinations in LLMs and their Mitigation\\
}

\author{
\IEEEauthorblockN{Parth Pulkundwar}
\IEEEauthorblockA{\textit{Computer Engineering} \\
K J Somaiya Institute of Technology \\
Mumbai, India \\
p.pulkundwar@somaiya.edu}
\and
\IEEEauthorblockN{Vivek Dhanawade}
\IEEEauthorblockA{\textit{Artificial Intelligence and Data Science} \\
K J Somaiya Institute of Technology \\
Mumbai, India \\
vivek.dhanawade@somaiya.edu}
\and
\IEEEauthorblockN{Rohit Yadav}
\IEEEauthorblockA{\textit{Information Technology} \\
K J Somaiya Institute of Technology \\
Mumbai, India \\
rohit15@somaiya.edu}
\and
\IEEEauthorblockN{Prof. Minal Sonkar}
\IEEEauthorblockA{\textit{Computer Engineering} \\
K J Somaiya Institute of Technology \\
Mumbai, India \\
minal.sonkar@somaiya.edu}
\and
\IEEEauthorblockN{Prof. Medha Asurlekar}
\IEEEauthorblockA{\textit{Artificial Intelligence and Data Science} \\
K J Somaiya Institute of Technology \\
Mumbai, India \\
medha@somaiya.edu}
\and
\IEEEauthorblockN{Prof. Sarita Rathod}
\IEEEauthorblockA{\textit{Information Technology} \\
K J Somaiya Institute of Technology \\
Mumbai, India \\
sarita.r@somaiya.edu}
}

\maketitle

\begin{abstract}
Traditional language models face a challenge from hallucinations. Their very presence casts a large, dangerous shadow over the promising realm of natural language processing. It becomes crucial to understand the various kinds of hallucinations that occur nowadays, their origins, and ways of reducing them. This document provides a concise and straightforward summary of that. It serves as a one-stop resource for a general understanding of hallucinations and how to mitigate them.\newline
\end{abstract}


\section{Introduction}
In the fast-moving world of Natural Language Processing (NLP) today, large language models (LLMs) such as GPT, BERT, and others have become the principal agents of change in natural language processing. They can generate human-like text, answer multifaceted questions, or engage in conversation with as much fluency. If a method produces an output that makes sense ex-ante, we can think of it as being near north on a compass. However, an answer generated by LLMs may be related to the input and follow logical arguments, but has no truth value since this method is not based on known facts, and therefore its plausible excuse is a type of mirage that cannot be debunked from ex-ante knowledge. This phenomenon is called hallucination, a word that sums up the main obstacle to its widespread use. \newline

In other words, think of it as an experienced storyteller narrating their tale. At times, the storyteller gets so carried away in his flow that he includes details outside of reality all while attempting to tell the truth. In the same way, LLMs may produce content that seems accurate but has no grounding in the actual data they were trained on. A simpler analogy would be a GPS giving directions confidently but occasionally leading you down a nonexistent road. \newline

More technically, hallucinations in LLMs take the form of generated content that does not match factual or expected outputs relative to their training data. It could be with false statements, references fabricated, or even with an incorrect reasoning path. Just like the misguided GPS in the earlier analogy, the LLM seems to know where it’s going, but ends up in the wrong place. \newline

This is a challenge that the AI community is currently addressing. Hallucinations can occur even in advanced models. Although LLMs are remarkable, they are not consistently reliable, especially when it comes to producing factual data, and in contexts like healthcare and legal matters, the stakes are significant. This also applies to academic research.  \newline

Addressing hallucinations in conversational agents like chatbots and virtual assistants is particularly important. These systems are increasingly integrated into customer service, education, and decision-making platforms, where users rely on the accuracy of the information they receive. Unchecked hallucinations could lead to misinformation, eroding trust in AI systems and limiting their potential applications. \newline

This review aims to delve into the nature of hallucinations in LLMs, investigate the current state of research, and explore potential solutions. By understanding the mechanisms behind hallucinations and the efforts to mitigate them, this review aims to highlight the importance of this challenge in ensuring the future reliability of AI-driven language systems. \newline

\section{Literature Review}

The problem of hallucinations in LLMs has become a hot topic, with several approaches targeting its understanding, detection, and mitigation features. One such early work is [1], which discusses hallucinations in NLP in great detail and attempts to define them based on their typology, as well as their motives and causes.  [5] is another work in this area, tackling the problem of hallucinations in LLMs and more specifically, the challenges one encounters in terms of their use in important contexts. Daugherty et al. [17] enhance this understanding by reviewing the problem of hallucination across most of the NLP tasks. Meanwhile [20] highlights the existing challenge of hallucinations in areas of application where precision is critical. In the equally fascinating, yet less optimistic paper [24], the author argues that hallucinations are simply part and parcel of LLMs, and therefore, mechanisms must be put in place to deal with this issue. \newline

It is imperative to address the hallucinations present in generated outputs. In this regard, it is also of paramount importance to make progress in the evaluation and detection of contradictory hallucinations.  [3] helps in the development of a tool to identify hallucinations, which [14] and [15], are both developing methods that do not require any resources for hallucination detection including the model's internal structure. It was theorized in [16] that LLMs are capable of knowing when they are submitting false information, and it is suggested how this ability can be used for detection. [22] uses unlabeled LLM completions for the detection of hallucinations while [12] aims at early detection within the question-answering systems to avoid the spread of false information. \newline

There has been development of fine-tuning and model adaptation methods to address hallucination effects. In [4] the authors consider data augmentation and changes of the model to lessen the number of hallucinations. Designers propose a fine-tuning technique that overcomes human learning process-related hallucination as used in large language models with an emphasis on cost efficiency. [25] deals with finessing the models in such a way that they are more true to the training dataset that was provided. While [9] focuses on external knowledge available that can be used to alleviate hallucinations. On the other hand, [26] looks into the effects of hallucinations when new knowledge is used for fine-tuning. [21] proposes an approach based on prompts guided by formal methods to lower the probability of hallucinations occurring. \newline

User-central approaches are equally significant in minimizing the effects of hallucinations. In [2] the authors emphasize the importance of interaction with the users in real-time that would allow users to spot and fix any hallucinations. [11] is an extensive review of bias mitigation and dehallucination methods in LLMs to increase their acceptance by humans. \newline

With retrieval-augmented and constraint-based methods, hallucinations can be reduced as well. [6] investigates how enhancing the constraints of generations can help diminish hallucinations without compromising fluency. [18] points out the need for external knowledge references to improve the precision of conversational LLMs. [8] is concerned with having retrieval augmentation incorporated within financial reporting tasks. Several strategies in [23] apply and advocate for the use of several different approaches with the aim of hallucinations but from different perspectives. \newline

Domain-specific approaches tailor hallucination mitigation strategies even further. [19] proposes some measures that one can take to lessen the incidence of hallucinations while generating educational content. [9] also advocates for knowledge injection to reduce hallucinations, especially in the voice of customer reviews of retail shops. [8] explains strategies for reducing hallucinations in information extraction from financial data, which is an important aspect due to the need for precision. These domain-specific strategies illustrate how enhancement techniques may be adapted for specific sectors, thus increasing confidence in LLMs. \newline

\section{Hallucinations in Language Models}
Explained in simpler terms, hallucinations of language models are major issues since they influence how credible the content generated is. Hallucinations are many: some are factual and logical. A factual hallucination is when the model provides a lie, whilst logical inconsistency is when the output is contradictory or incoherent with itself. Such hallucinations will often lower the user's trust towards the system, but domains such as, health care and finance, where critical applications exist may not cope with such hallucinations.

Some approaches have also been provided to hallucination detection in LLM outputs and focus on both factual and logical errors. [13] addresses the issue of detection of one particular form of logical error that is typically of great interest – the errors in the form of contradictory statements made by the model. The paper provides the obstacles and the potential for addressing these deceptively beguiling contradictions suggesting a way to increase the trustworthiness of LLMs. [3] is a tool that helps detect hallucinations by examining the responses of the model for coherence and factuality. This tool, however, aims at correcting the gap that exists between the generated content and what is already known to be true.

Building upon the foundations laid by the previous two works, [14] proposes the novel detection of hallucination without the need for model internals, or internal core architecture access. This zero-access model is most useful in practical situations, for instance, in hallucination detection, when no access to internal model details is possible. In addition, [15] investigates if it is possible to discern those which are hallucinated without fine-tuning the models or searching for the relevant concepts externally.

Another instance is that [16], which notes that hallucinations, caused by the fabrications - in these cases - may be monitored by LLMs self-assessing their internal states. The upside of this is the potential for self-aware agents where the AI can detect untrustworthy outputs. [22], suggests using LLM decoded text that is not labeled already instead of precluding labeled data. This aids in building hallucination detection models, which will ease the process significantly by reducing the amount of man-hours physically spent to train the models.

Last but not least, [12] discusses the significance of the question-answering system with extensive underlying research and real-time strategy on hallucinations of concrete information to avert the spread of wrong ideas. This is especially important for the end-users of LLMs, which possess generative capabilities and the possibility of interaction with a user.

All in all, these papers are aimed at finding ways to diminish the hallucinations in the text generation, while guaranteeing more credibility and trustworthiness of the content produced by the LLMs. There is great potential for strategies such as zero-resource detection as well as self-awareness and early detection in combating hallucinations in language models owing to the various difficulties they present.

\begin{figure}
    \centering
    \includegraphics[width=1\linewidth]{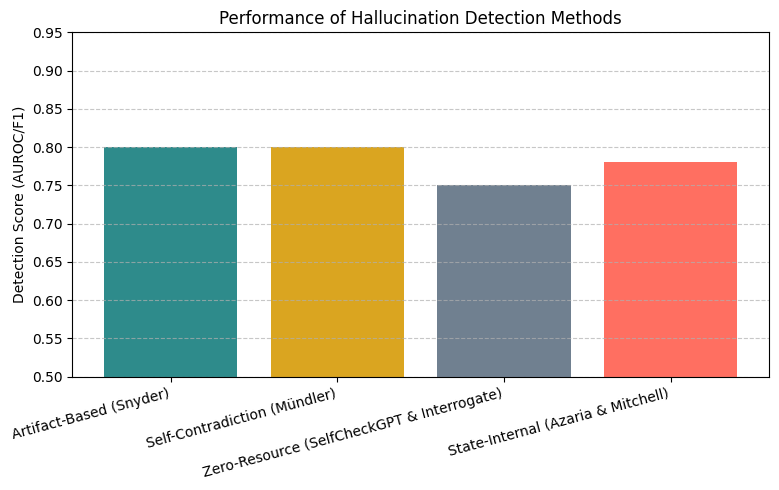}
    \caption{ Hallucination In Language Models}
    \label{fig:enter-label}
\end{figure}

\section{LLMs and Hallucinations}
It is important to be cognizant of the link between LLMs and hallucinations, as this influences the choice of the models that mitigate this problem. Several and quite distinct LLMs differ in the degree to which they produce hallucinations, often due to the differences in the architecture of the model and the training process. \newline

The paper [1] provides a comprehensive discussion on hallucinations in language models, outlining the discomfort experienced by the models while producing illogical and non-factual outputs. It also stresses the need to comprehend the rationale and classes of hallucinations incurred by NLP systems including fabrication of facts and meaningless expression. Likewise, [5] explains the reasons why these models are active hallucination constructors, due to the size of the training dataset and its complexity, at times presenting inaccurate states or even creating information that does not exist. \newline

Paper [17] discusses various hallucination types, classified into many thematic groups, broadening on why some LLMs may be more reliable than others. This review addresses the influence of model size, structure, and training data in the occurrence of these hallucinations. [20] presents these tendencies across different models and offers solutions like improvement of training techniques and providing limitations in the generated output to tackle the issue. \newline

In the same breath, [24], notes that hallucination remains a fundamental problem of LLMs, regardless of their dimensions or training methodology. Propose that it is possible to restrain hallucinations by a thorough model selection and training, however, the authors do not believe that such models can be inducted completely free from hallucinations. This is of great significance as it helps with formulating realistic goals and the appropriateness of the chosen model for the given task. \newline

In choosing an LLM to reduce hallucinations: model architecture, quality and diversity of the training data, and strategies like fine-tuning or retrieval augmentation to alleviate hallucinations should be kept in mind. Many such models are available in the market these days such as GPT, BERT, T5, etc. Out of these, it has been observed that GPT often produces text that is logically and cohesively consistent with the content, though such generative models come at the disadvantage of several fabricated data (hallucinations). Such hallucinations tend to occur at lower rates for BERT due to the use of bidirectional training for a specific range of tasks; however, it could be affixed to creating new material. Several other models have been introduced such as PaLM as well as LLaMa about hallucination reduction, the extent to which they succeed varies based on their training methods and designs. Knowing the advantages and limitations of such models plays a critical role when choosing a given tool to be used in applications where the accuracy of the information provided is paramount and hallucination rates have to be kept at a minimum.
 \newline

\section{Methodologies to Combat Hallucination}

There has been an increasing desire to minimize hallucinations in LLMs, leading to the development of several techniques that seek to alter the training of such systems, access outside knowledge, and even modify prompt generation. Such approaches strive to ensure that hallucinations become less frequent and less serious, as a result enabling better and safer usage of LLMs.

\textbf{Tuning and Retraining:} A common initiative aimed at diminishing the level of hallucinations is tuning models on data from the specific domain. The paper [4] emphasizes the need to modify models by retraining them to better cognize specific factual aspects. However, this approach comes with the caveat that the primary purpose of the training data should not be to clean prevailing traces of rot or hallucination promotion tendencies. In the same way, [10] explains how a new fine-tuning technology inspired by humanity’s ways of learning can achieve a hallucination-free large language model while utilizing limited processing capabilities. \newline

\textbf{Reinforcement Learning (RL):} One technique that can be leveraged to address hallucination is Reinforcement Learning from Human Feedback (RLHF). RLHF has found its application in improving the alignment of large language models (LLMs) towards human preferences, but its use to reduce hallucination is more complicated. In [25], the authors provide a deep analysis of its positive applications, particularly RLHF in promoting what they regard as good behavior, and in this case, concerning the fidelity of the outputs in terms of their truthfulness. Therefore, the fine-tuning of RL models is pertinent as it assists in democratizing the generative actions of the models in line with what is supposed to be output to the users, and as such, eliminating distortions or hallucinations in the content produced. \newline

\begin{table}[ht]
\centering
\caption{RL-Based Hallucination Mitigation Methods}
\label{tab:rlhf_methods}
\begin{tabular}{@{}p{1.5cm}p{1.8cm}p{1.8cm}p{1.5cm}@{}}
\toprule
\textbf{Method} & \textbf{Domain / Task} & \textbf{Human/Dense Feedback} & \textbf{Key Metric} \\
\midrule
RLHF & General LLM alignment & Human pairwise preferences & Improved factual behavior \\
Fact-RLHF & Vision--language (LMMs) & Human annotation + factual augment. & 96\% vs 87\% baseline accuracy [2] \\
RLHF-V & Multimodal LLMs & Segment-level fine-grained human feedback (1.4K samples) & 34.8\% hallucination reduction [3] \\
\bottomrule
\end{tabular}
\end{table}

\textbf{Knowledge Integration:} Another potent weapon in the fight against hallucinations is assimilating external knowledge bases. [9] deals with the importance of how structured knowledge sources like databases or factual repositories within content are embedded into LLM to help them produce better and less fallacious content with the help of reliable information. Yet, [26] is more optimistic about the prospects and poses the problem of whether new knowledge made available within the modeled information might not create hallucinatory effects if any fine-tuning is introduced. The study advocates that the management of this process should be conducted without causing conflict between the new knowledge and the old knowledge in the model to avoid contradictory knowledge production outputs. \newline

\textbf{Comparative Methodologies:} Another method of addressing hallucinations is to assess different model fine-tuning optimization techniques. [21] puts forward formal methods of guided iterative prompting where the generation process is supported by some structure and rules thereby very little room is left for hallucinations. This approach is advantageous compared to standard fine-tuning as the content generation process can get iterations based on logical constraints. \newline

\textbf{Best Practices in Prompt Design and Post-Processing:} One of the ways of doing away with hallucinations entails the proper design of prompts and post-processing steps. A good prompt lessens the tendency of hallucinatory response by helping the model find the most factual and relevant answers. Such techniques as iterative prompting which helps steer the model away from response areas likely to contain a hallucination, or using constraints to inform prompts as is outlined in the techniques of [21]. Improvement of this restriction is also possible by designing post-processing tools that either block the output of unproven assertions or incorporate external fact-checking processes. \newline

\begin{figure}
    \centering
    \includegraphics[width=1\linewidth]{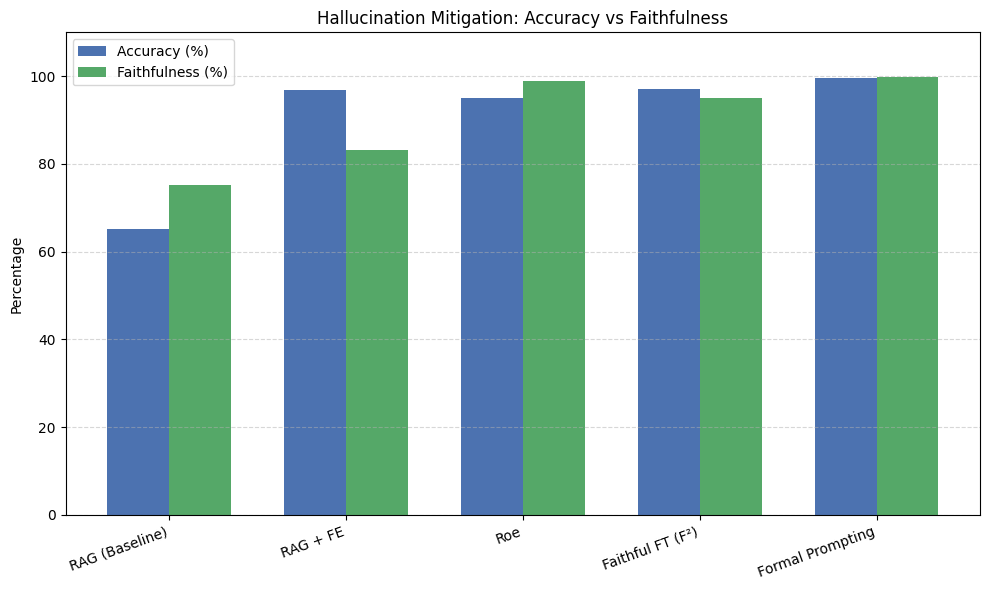}
    \caption{\textit{Comparison of Hallucination-Mitigation Techniques Across Key Metrics.} This bar chart presents \textbf{accuracy} (blue bars) and \textbf{faithfulness} (green bars) for five methods}
    \label{fig:enter-label}
\end{figure}

\begin{table}[ht]
\centering
\caption{Error Category Breakdown by Method}
\label{tab:error_breakdown}
\begin{tabular}{@{}lccc@{}}
\toprule
Method            & Factual Errors\footnotesize{(\%)} & Uncertain\newline Refusals\footnotesize{(\%)} & Other Errors\footnotesize{(\%)} \\
\midrule
RAG + FE         & 15.0                            & 2.0                              & 83.0 \\
Roe              & 5.0                             & 10.0                             & 85.0 \\
Faithful FT (F²) & 7.0                             & 3.0                              & 90.0 \\
Formal Prompting & 1.0                             & 0.0                              & 99.0 \\
\bottomrule
\end{tabular}
\end{table}

In summation, the amalgamation of fine-tuning, reinforcement learning, knowledge embedding, and effective prompt engineering is a strategic way of dealing with hallucinations in large language models. Every approach significantly enhances the factual reliability and the trustworthiness of such models for practical uses. \newline

\section{Evaluation and Metrics}

Preparing solid benchmarks, techniques, and metrics for hallucinations entails numerous activities that evaluate the extent to which models protect themselves from generating untrue or otherwise deceptive content. Evaluation frameworks and associated methods of this kind are already available in some variants: herein we focus on those that pertain to hallucination detection, model performance assessment, and relative evaluation of mitigation frameworks' effectiveness. \newline

\textbf{Existing Benchmarks for Hallucination Detection:} Several benchmarks have been developed to quantify the reviewed hallucinations within LLM-generated outputs. TruthfulQA is a representative example that aims at evaluating LLM’s performance based on the veracity of its answers. Besides, specifically for dialogue-based models, FaithDial gauges the performance by looking at how consistently and truthfully a model can respond throughout the conversation. These benchmarks provide the effectiveness of the models at mitigating hallucinations as well as ensuring that no falsehoods are presented in the output. \newline

In the article [1], the authors analyze approaches to hallucination detection, in particular the necessity of structured benchmarks that assess the factuality and logical consistency of the models. They have also noted the existing difference between the way hallucination is understood by humans and the output generated by the model indicating that sophisticated metrics should be applied in such measurements. \newline

\textbf{Evaluation Metrics and Techniques:} To analyze the performance of the Language Model concerning hallucination mitigation, multiple detection techniques have been put to use. [3] suggests a device that detects hallucinations in pieces of text produced by the Language Model with the help of fact-checking. This is taken a step further by offering a black box model that is capable of detecting hallucinations using external means even when the inner workings do not have access to the model's architecture. This and other approaches help support pragmatic estimates of the hallucinations contained in LLMs, which is vital in many deployments. [15] also presents a non-intrusive detection system ‘in the wild’ aiming at spotting inconsistencies and hallucinations of LLM responses without studying the model or additional resources management. \newline

\textbf{User-Centered Evaluations:} Because hallucinations affect the user's level of trust in the system, considerations of how the user experiences these errors become very important. [2] provides an evaluation of hallucinations in the context of language models, in which users actively engage in hallucination detection and reporting. This strategy encompasses new user-oriented metrics that include the impact of hallucinations - loss of trust in the system and the burden of self-verification onerous to the users which the system is supposed to take over. \newline

\textbf{Evaluation of Hallucinations: }A More Inclusive Framework for Hallucination Evaluation: Hallucinations and their effects are assessed and controlled using a range of measures that have contributed to the establishment of more integrated frameworks. [23] illustrates that performance evaluations of LLMs can be done comprehensively by employing several evaluation techniques including factuality checks, consistency analysis, and user experience evaluations. This broad-based approach helps to identify hallucinations from various perspectives, looking not only at the rightness or wrongness of facts but also at the soundness of the arguments and the user experience. \newline

\section{Recommendations}
\textbf{Applications of LLMs in Different Scenarios:}
While it is essential to use large language models (LLMs) in any application that requires understanding a task or situation without considering the models' propensity for hallucination, their applicability is dependent also on the carnality of the task. As far as general tasks such as chatbots are concerned, integrating models such as GPT-4 or PaLM is common because it can create fluent, human-like writing with little risk of veracity, and infuses just the right amount of creativity. However, applications involving systems that require factual answers such as retrieval or question answering are most likely BERT and T5-based due to the models' grounded understanding and dependency on facts. \newline

With specialized knowledge (e.g. medical or financial domains), hallucinations can be further controlled by training LLaMA and other models using domain-specific datasets. Besides, Retrieval-augmented Generation (RAG) models have also been able to mitigate hallucinations, as they are capable of incorporating external information that has already been validated. Thus, these models are ideal in changing circumstances, especially those that demand meticulous fidelity to facts. \newline

\textbf{Optimal Strategies for Alleviating Hallucinations in Language Models:}
There are various techniques to address the problem of hallucinations in LLMs, which will depend on the type of methodology chosen, application in question, and resource availability. A common approach is referred to as fine-tuning in which LLMs undergo training with a narrow corpus of data and enables the models to produce outputs that are more consistent with facts. For instance, as shown in [25], it is possible to enforce enhancement of the factual reliability of large language models when they are fine-tuned on high-quality curated datasets. A different approach that is additional knowledge integration is prevalent in use today as well. Incorporating or attaching external knowledge bases in the model via RAG technique makes it less prone to hallucinations by ensuring that all of the model's replies are backed up by some credible sources. This method has advantages in areas that deal with current information or very specific information, as discussed in [8]. \newline

\textbf{Reinforcement Learning from Human Feedback (RLHF): }
RLHF is also becoming popular as an effective approach for ensuring that the outputs of LLMs meet human expectations, reducing the incidence of hallucinations, and more generally, improving the quality of content generation. By refining the models with feedback from users, they manage to learn the importance of responding truthfully, accurately, and consistently, as explained in [10]. \newline

\textbf{Future Research Directions and Advancements:}
At present, the phenomenon of hallucination in LLMs is still a major concern, which means many aspects can be considered in future research and are rather attractive. One of them would be the real-time systems for the detection of hallucination, which would be responsible for identifying, and more importantly correcting, such contrary to reality outputs during the model’s inference rather than the correction being done afterward. This may further improve the view of LLMs within their deployment in real-world situations that involve high risks such as generating legal or medical documents whose inaccuracy might be harmful or quite catastrophic. \newline

Another aspect that is being looked into is the multi-modal detection of hallucinations which is, apart from text, other kinds of data (images, audio or structured data) are considered to enhance the model’s performance and this is also important with most, if not all LLMs being used in multi-modal contexts where hallucinations can occur due to the combination of texts with other forms of media. \newline

User-centered approaches are now of interest where models have features that help users to easily spot, rectify or erase the hallucinations. Conducting studies that are aimed at the invention of interactive and transparent AI systems that users can interact with by inquiring and confirming what the model has said is an interesting area of research, which has been discussed in detail in [2]. Lastly, there is a crucial need to conduct more studies about hallucination mitigation using formal methods, as indicated in [21], which is certain to offer a model reliability guarantee techniques that are based on the formal methods approach useful in cases that have stringent accuracy requirements. \newline

\section{Conclusion}

This article reviewed some of the hallucinations found among LLMs, classifying those hallucinations into sub-classes such as distortion of facts or logic resulting from scarcity of information, ambiguity of requested inputs, or even limitations of the model. Hallucinations lower the trust of end users in the system and the usability of the system as well hence calling for strategies on how to reduce it. \newline

It is regarding the concept of reducing these hallucinations mainly because of their negative effects, that the following approaches; domain adaptation, knowledge bases, and reward-based training, have been promising. To meet the challenges of hallucinations, retrieval augmented generation (RAG) and real-time detection mechanisms have been identified as helpful in enhancing the accuracies of the models. TruthfulQA and FaithDial are examples of targeted evaluation metrics that can be deployed to test the factuality of the outputs produced by the models. \newline

Creating reliable AI systems does not remain static; thus, these strategies need to be developed further. Research in the coming years should concentrate on enhancing the detection and reduction of such biases so that harmful ideas and incorrect facts do not make it into the outputs of the AI system. The continuous development of these techniques will be necessary to relieve concerns that may be raised during the deployment of LLMs.

\end{document}